\documentclass{article}

\PassOptionsToPackage{sort,numbers}{natbib}
\usepackage{natbib}

\usepackage{comment}
\usepackage{times}
\usepackage{latexsym}
\usepackage{multirow}
\usepackage{tabularx}

\usepackage{floatrow}

\usepackage[percent]{overpic}
\usepackage{url}

\usepackage{pifont}
\newcommand{\cmark}{\ding{51}}%
\newcommand{\xmark}{\ding{55}}%
\usepackage{tikz}
\usepackage{amsmath}
\usepackage{amssymb}
\usepackage{mathabx}
\usepackage{enumitem}
\usetikzlibrary{shapes,arrows}
\usetikzlibrary{positioning}
\usepackage{subcaption}

\usepackage[final]{nips_2018}
\usepackage{graphicx}

\usepackage{array}
\usepackage{pgfplots}
\usepackage{pgfplotstable}
\usetikzlibrary{fit,positioning,arrows}

\usepackage[utf8]{inputenc} 
\usepackage[T1]{fontenc}    
\usepackage{hyperref}       
\usepackage{url}            
\usepackage{booktabs}       
\usepackage{amsfonts}       
\usepackage{nicefrac}       
\usepackage{microtype}      
\usepackage{xfrac}          
\usepackage{color}          
\usepackage{xcolor}          
\usepackage[]{algorithm2e}
\usepackage{algorithmicx}
\usepackage{algpseudocode}
\usepackage{float}
\usepackage{wrapfig}
\restylefloat{figure}

\newcommand{\cuttablebelow}{\vspace*{-0.1in}}

\usepackage{nips_2018}

\usepackage[utf8]{inputenc} 
\usepackage[T1]{fontenc}    
\usepackage{hyperref}       
\usepackage{url}            
\usepackage{booktabs}       
\usepackage{amsfonts}       
\usepackage{nicefrac}       
\usepackage{microtype}      

\usetikzlibrary{%
  arrows,%
  shapes.misc,
  shapes.arrows,%
  chains,%
  matrix,%
  positioning,
  scopes,%
  decorations.pathmorphing,
  shadows%
}

\title{Content preserving text generation \\ with attribute controls}

\author{
  Lajanugen Logeswaran$^*$ \hspace{1em} Honglak Lee$^{\dagger}$ \hspace{1em} Samy Bengio$^{\dagger}$ \\
  $^*$University of Michigan, $^\dagger$Google Brain  \\
  \texttt{llajan@umich.edu,\{honglak,bengio\}@google.com}
}

\begin{document}

\maketitle

\begin{abstract}
\vspace{-0.05in}


In this work, we address the problem of modifying textual attributes of sentences. 
Given an input sentence and a set of attribute labels, we attempt to generate sentences that are compatible with the conditioning information.
To ensure that the model generates content compatible sentences, we introduce a reconstruction loss which interpolates between auto-encoding and back-translation loss components. 
We propose an adversarial loss to enforce generated samples to be attribute compatible and realistic.
Through quantitative, qualitative and human evaluations we demonstrate that our model is capable of generating fluent sentences that better reflect the conditioning information compared to prior methods.
We further demonstrate that the model is capable of simultaneously controlling multiple attributes.
\vspace{-0.2in}
\end{abstract}

\section{Introduction}

Generative modeling of images and text has seen increasing progress over the last few years. 
Deep generative models such as variational auto-encoders \citep{kingma2013auto}, adversarial networks \citep{goodfellow2014generative} and Pixel Recurrent Neural Nets \citep{oord2016pixel} have driven most of this success in vision. 
Conditional generative models capable of providing fine-grained control over the attributes of a generated image such as facial attributes \citep{yan2016attribute2image} and attributes of birds and flowers \citep{reed2016generative} have been extensively studied.
The style transfer problem which aims to change more abstract properties of an image has seen significant advances \citep{gatys2015neural,isola2016image}.


The discrete and sequential nature of language makes it difficult to approach language problems in a similar manner.
Changing the value of a pixel by a small amount has negligible perceptual effect on an image.
However, distortions to text are not imperceptible in a similar way and this has largely prevented the transfer of these methods to text.


In this work we consider a generative model for sentences that is capable of expressing a given sentence in a form that is compatible with a given set of conditioning attributes.
Applications of such models include conversational systems \citep{li2016persona}, paraphrasing \citep{xu2012paraphrasing}, machine translation \citep{sennrich2016controlling}, authorship obfuscation \citep{shetty2017author} and many others.
Sequence mapping problems have been addressed successfully with the sequence-to-sequence paradigm \citep{sutskever2014sequence}.
However, this approach requires training pairs of source and target sentences.
The lack of parallel data with pairs of similar sentences that differ along certain stylistic dimensions makes this an important and challenging problem.

We focus on categorical attributes of language. 
Examples of such attributes include sentiment, language complexity, tense, voice, honorifics, mood, etc. 
%
%
%
%
Our approach draws inspiration from style transfer methods in the vision and language literature.
We enforce content preservation using auto-encoding and back-translation losses.
Attribute compatibility and realistic sequence generation are encouraged by an adversarial discriminator.
The proposed adversarial discriminator is more data efficient and scales better to multiple attributes with several classes more easily than prior methods.

Evaluating models that address the transfer task is also quite challenging. 
Previous works have mostly focused on assessing the attribute compatibility of generated sentences. 
These evaluations do not penalize vacuous mappings that simply generate a sentence of the desired attribute value while ignoring the content of the input sentence. 
This calls for new metrics to objectively evaluate models for content preservation. 
In addition to evaluating attribute compatibility, we consider new metrics for content preservation and generation fluency, and evaluate models using these metrics. 
We also perform a human evaluation to assess the performance of models along these dimensions.

We also take a step forward and consider a writing style transfer task for which parallel data is available. 
Evaluating the model on parallel data assesses it in terms of all properties of interest: generating content and attribute compatible, realistic sentences.
Finally, we show that the model is able to learn to control multiple attributes simultaneously.
To our knowledge, we demonstrate the first instance of learning to modify multiple textual attributes of a given sentence without parallel data.

\section{Related Work}
\textbf{Conditional Text Generation}
\, Prior work have considered controlling aspects of generated sentences in machine translation such as length \citep{kikuchi2016controlling}, voice \citep{yamagishi2016controlling}, and honorifics/politeness \citep{sennrich2016controlling}.
\citet{kiros2014multiplicative} use multiplicative interactions between a word embeddings matrix and learnable attribute vectors for attribute conditional language modeling. 
\citet{radford2017learning} train a character-level language model on Amazon reviews using LSTMs \citep{hochreiter1997long} and discover that the LSTM learns a `sentiment unit'.
By clamping this unit to a fixed value, they are able to generate label conditional paragraphs.

\citet{hu2017controllable} propose a generative model of sentences which can be conditioned on a sentence and attribute labels.
The model has a VAE backbone which attempts to express holistic sentence properties in its latent variable. 
A generator reconstructs the sentence conditioned on the latent variable and the conditioning attribute labels.
Discriminators are used to ensure attribute compatibility.
Training sequential VAE models has proven to be very challenging \citep{bowman2015generating,chen2016variational} because of the posterior collapse problem.
Annealing techniques are generally used to address this issue. 
However, reconstructions from these models tend to differ from the input sentence.

\textbf{Style Transfer}
\, Recent approaches have proposed neural models learned from non-parallel text to address the text style transfer problem.
\citet{li2018delete} propose a simple approach to perform sentiment transfer and generate stylized image captions. 
Words that capture the stylistic properties of a given sentence are identified and masked out, and the model attempts to reconstruct the sentence using the masked version and its style information.
\citet{shen2017style} employ adversarial discriminators to match the distribution of decoder hidden state trajectories corresponding to real and synthetic sentences specific to a certain style.
\citet{prabhumoye2018style} assume that translating a sentence to a different language alters the stylistic properties of a sentence.
They adopt an adversarial training approach similar to \citet{shen2017style} and replace the input sentence using a back-translated sentence obtained using a machine-translation system.


To encourage generated sentences to match the conditioning stylistic attributes,  prior discriminator based approaches train a classifier or adversarial discriminator specific to each attribute or attribute value. 
In contrast, our proposed adversarial loss involves learning a single discriminator which determines whether a sentence is both realistic and is compatible with a given set of attribute values. 
We demonstrate that the model can handle multiple attributes simultaneously, while prior work has mostly focused on one or two attributes, which limits their practical applicability.

\textbf{Unsupervised Machine Translation}
\, There is growing interest in discovering latent alignments between text from multiple languages.
Back-translation is an idea that is commonly used in this context where mapping from a source domain to a target domain and then mapping it back should produce an identical sentence. 
\citet{he2016dual} attempt to use monolingual corpora for machine translation. 
They learn a pair of translation models, one in each direction, and the model is trained via policy gradients using reward signals coming from pre-trained language models and a back-translation constraint.
\citet{artetxe2017unsupervised} proposed a sequence-to-sequence model with a shared encoder, trained using a de-noising auto-encoding objective and an iterative back-translation based training process.
\citet{lample2017unsupervised} adopt a similar approach but with an unshared encoder-decoder pair.
In addition to de-noising and back-translation losses, adversarial losses are introduced to learn a shared embedding space, similar to the aligned-autoencoder of \citet{shen2017style}.
While the auto-encoding loss and back-translation loss have been used to encourage content preservation in prior work, we identify shortcomings with these individual losses: auto-encoding prefers the copy solution and back-translated samples can be noisy or incorrect. 
We propose a reconstruction loss which interpolates between these two losses to reduce the sensitivity of the model to these issues.





\vspace{-0.5em}
\section{Formulation}
\begin{figure}
\begin{tikzpicture}[
    >=latex,thick,
    /pgf/every decoration/.style={/tikz/sharp corners},
    node distance=4mm
  ]

    \begin{scope}[start chain,
            every node/.style={on chain},
        ]
        \node (input)       {\small I will go to the airport .};
        \node [xshift=2mm,inner sep=0pt,minimum size=0pt]  (after input)      {};
    \end{scope}
    \node (A) [above right=0.18cm and 5.9cm of input] {\small I went to the airport.};
    \node (B) [below right=0.22cm and 5.9cm of input] {\small I would go to the airport.};
    \node (C) [right=5.9cm of input] {\small I am going to the airport.};
    \begin{scope}[->,decoration={post length=0pt},rounded corners=2mm]
        \draw[-] (input) -- (after input);
        \draw (after input) |- node[anchor=center,above right=0.001cm and 0.1cm] {\small \textit{mood=indicative, tense=past}} (A);
        \draw (after input) |- node[anchor=center,above right=0.001cm and 0.1cm] {\small \textit{mood=subjunctive, tense=conditional}} (B);
        \draw (after input) -- node[anchor=center,above right=0.001cm and -2.55cm] {\small \textit{mood=indicative, tense=present}} (C);
    \end{scope}
\end{tikzpicture}
\caption{Task formulation - Given an input sentence and attributes values (Eg: indicative mood, past tense) generate a sentence that preserves the content of the input sentence and is compatible with the attribute values.  }
\label{fig:task}
\end{figure}
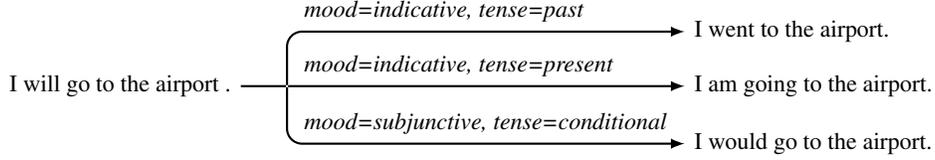
%
%
Suppose we have $K$ attributes of interest $\{a_1,...,a_{K}\}$.
Let be given a set of labelled sentences $D = \{(x^n, l^n) \}_{n=1}^N$ where $l^n$ is a set of labels for a subset of the attributes.
Given a sentence $x$ and attribute values $l'=(l_1,...,l_{K})$ our goal is to produce a sentence that shares the content of $x$, but reflects the attribute values specified by $l'$ (figure \ref{fig:task}).
In this context, we define \textit{content} as the information in the sentence that is not captured by the attributes.
We use the term \textit{attribute vector} to refer to a binary vector representation of the attribute labels. 
This is a concatenation of one-hot vector representations of the attribute labels.

\subsection{Model Overview}


We denote the generative model by $G$.
We want $G$ to use the conditioning information effectively.
i.e., $G$ should generate a sentence that is closely related in meaning to the input sentence and is consistent with the attributes.
We design $G = (G_\text{enc}, G_\text{dec})$ as an encoder-decoder model.
The encoder is an RNN that takes the words of input sentence $x$ as input and produces a content representation $z_x = G_\text{enc}(x)$ of the sentence.
Given a set of attribute values $l'$, a decoder RNN generates sequence $y \sim p_G(\cdot|z_x,l')$ conditioned on $z_x$ and $l'$.

\subsection{Content compatibility}
\label{sec:content}

We consider two types of reconstruction losses to encourage content compatibility.

\textbf{Autoencoding loss}
\, Let $x$ be a sentence and the corresponding attribute vector be $l$.
Let $z_x = G_\text{enc}(x)$ be the encoded representation of $x$.
Since sentence $x$ should have high probability under $G(\cdot|z_x,l)$, we enforce this constraint using an auto-encoding loss.
\begin{equation}
    \mathcal{L}^{ae} (x,l) = -\text{log } p_G(x|z_x, l)
\end{equation}
  

\textbf{Back-translation loss}
\, Consider $l'$, an arbitrary attribute vector different from $l$ (i.e., corresponds to a different set of attribute values).
Let $y \sim p_G(\cdot | z_x, l')$ be a generated sentence conditioned on $x, l'$.
Assuming a well-trained model, the sampled sentence $y$ will preserve the content of $x$.
In this case, sentence $x$ should have high probability under $p_G(\cdot|z_y,l)$ where $z_y = G_\text{enc}(y)$ is the encoded representation of sentence $y$.
This requirement can be enforced in a back-translation loss as follows.
\begin{equation}
    \mathcal{L}^{bt} (x,l) = -\text{log } p_G(x|z_y, l)
\end{equation}

A common pitfall of the auto-encoding loss in auto-regressive models is that the model learns to simply copy the input sequence without capturing any informative features in the latent representation.
A de-noising formulation is often considered where noise is introduced to the input sequence by deleting, swapping or re-arranging words.
On the other hand, the generated sample $y$ can be mismatched in content from $x$ during the early stages of training, so that the back-translation loss can potentially misguide the generator.
We address these issues by interpolating the latent representations of ground truth sentence $x$ and generated sentence $y$.

\textbf{Interpolated reconstruction loss}
\, We merge the autoencoding and back-translation losses by fusing the two latent representations $z_x,z_y$.
We consider $z_{xy} = g \odot z_x + (1 - g) \odot z_y$, 
where $g$ is a binary random vector of values sampled from a Bernoulli distribution with parameter $\Gamma$.
We define a new reconstruction loss which uses $z_{xy}$ to reconstruct the original sentence.
\begin{equation}
   \mathcal{L}^{int} = \mathbb{E}_{(x,l)\sim p_\text{data}, y \sim p_G(\cdot | z_x,l')} [-\text{log } p_G(x|z_{xy},l)]
\end{equation}

Note that $\mathcal{L}^\text{int}$ degenerates to $\mathcal{L}^\text{ae}$ when $g_i = 1$ $\forall i$, and to $\mathcal{L}^\text{bt}$ when $g_i = 0$ $\forall i$.
The interpolated content embedding makes it harder for the decoder to learn trivial solutions since it cannot rely on the original sentence alone to perform the reconstruction.
Furthermore, it also implicitly encourages the content representations $z_x$ and $z_y$ of $x, y$ to be similar, which is a favorable property of the encoder.

\begin{figure*}[!t]
    \centering
    \includegraphics[width=1.0\textwidth,trim={0 13cm 5.0cm 0.3cm},clip]{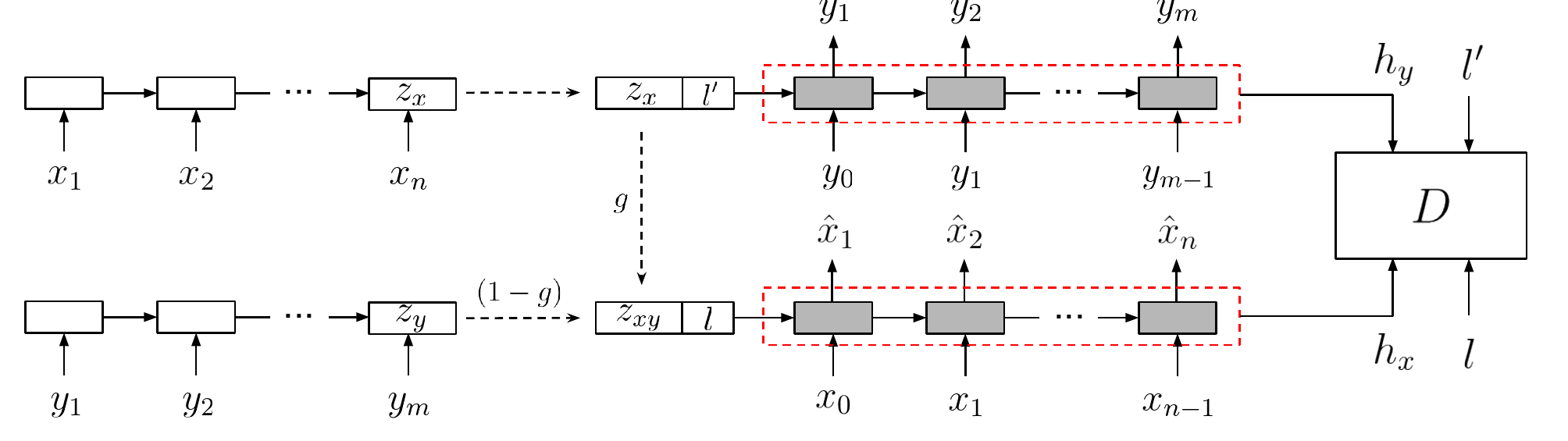}
    \caption{Given an input sentence $x$ with attribute labels $l$, we construct an arbitrary label assignment $l'$ and sample a sentence $y \sim p_G(x,l')$. Given the content representations of $x$ and $y$, an interpolated representation $z_{xy}$ is computed. The decoder reconstructs the input sentence using $z_{xy}$ and $l$. An adversarial discriminator $D$ encourages sequence $y$ to be both realistic and compatible with $l'$. \vspace{-1em} }
    \label{interp}
\end{figure*}
                                                             
\subsection{Attribute compatibility}

We consider an adversarial loss which encourages generating realistic and attribute compatible sentences.
The advesarial loss tries to match the distribution of sentence and attribute vector pairs $(s,a)$ where the sentence can either be a real or generated sentence.
Let $h_x$ and $h_y$ be the decoder hidden-state sequences corresponding to $x$ and $y$ respectively.
We consider an adversarial loss of the following form, where $D$ is a discriminator.
Sequence $h_x$ is held constant and $l' \neq l$.
\begin{equation}
\mathcal{L}^\text{adv} = \min_G \max_D \mathbb{E}_{(x,l)\sim p_\text{data}, y \sim p_G(\cdot | z_x,l')} [\text{log} D(h_x, l) + \text{log}(1 - D(h_y, l'))]
\end{equation}
It is possible that the discriminator ignores the attributes and makes the real/fake decision based on just the hidden states, or vice versa.
To prevent this situation, we consider additional fake pairs $(x,l')$ similar to \cite{reed2016generative} where we consider a real sentence and a mismatched attribute vector, and encourage the discriminator to classify these pairs as fake.
The new objective takes the following form. 
\begin{equation}
\mathcal{L}^\text{adv} = \min_G \max_D \mathbb{E}_{\begin{subarray}{l}(x,l) \sim p_\text{data}\\y \sim p_G(\cdot|z_x, l')\end{subarray}} [2 \text{ log} D(h_x, l) + \text{log}(1 - D(h_y, l')) + \text{log}(1 - D(h_x, l'))]]
\end{equation}                                                                                                               
Our discriminator architecture follows the projection discriminator \citep{miyato2018cgans},
\begin{equation}
D(s, l) = \sigma(l_v^T W \phi(s) + v^T \phi(s))
\end{equation}
where $l_v$ represents the binary attribute vector corresponding to $l$. $\phi$ is a bi-directional RNN encoder ($\phi(\cdot)$ represents the final hidden state).
$W, v$ are learnable parameters and $\sigma$ is the sigmoid function. 

The overall loss function is given by $\mathcal{L}^\text{int} + \lambda \mathcal{L}^\text{adv}$ where $\lambda$ is a hyperparameter.

\subsection{Discussion}
\label{sec:discussion}

\textbf{Soft-sampling and hard-sampling} 
\, A challenging aspect of text generation models is dealing with the discrete nature of language, which makes it difficult to generate a sequence and then obtain a learning signal based on it.
Soft-sampling is generally used to back-propagate gradients through the sampling process where an approximation of the sampled word vector at every time-step is used as the input for the next time-step \citep{shen2017style,hu2017controllable}.
Inference performs hard-sampling, where sampled words are used instead.
Thus, when soft-sampled sequences are used at training time, the training and inference behavior are mismatched.
For instance, \citet{shen2017style}'s adversarial loss encourages the hidden-state dynamics of teacher-forced and soft-sampled sequences to be similar.
However, there remains a gap between the dynamics of these sequences and sequences hard-sampled at test time.
We eliminate this gap by hard-sampling the sequence $y$.
Soft-sampling also has a tendency to introduce artifacts during generation.
These approximations further become poor with large vocabulary sizes.
We present an ablative experiment comparing these two sampling strategies in Appendix \ref{sec:sampling}.

\textbf{Scalability to multiple attributes} 
\, \citet{shen2017style} use multiple class-specific discriminators to match the class conditional distributions of sentences.
In contrast, our proposed discriminator models the joint distribution of realistic sentences and corresponding attribute labels.
Our approach is more data-efficient and exploits the correlation between different attributes as well as attributes values.

\section{Experiments}

The sentiment attribute has been widely considered in previous work \citep{hu2017controllable,shen2017style}.
We first address the sentiment control task and perform a comprehensive comparison against previous methods.
We perform quantitative, qualitative and human evaluations to compare sentences generated by different models.
Next we evaluate the model in a setting where parallel data is available.
Finally we consider the more challenging setting of controlling multiple attributes simultaneously and show that our model easily extends to the multiple attribute scenario.


\subsection{Training and hyperparameters}
\label{hyperparameters}
We use the following validation metrics for model selection.
The autoencoding loss $\mathcal{L}^\text{ae}$ is used to measure how well the model generates content compatible sentences.
Attribute compatibility is measured by generating sentences conditioned a set of labels, and using pre-trained attribute classifiers to measure how well the samples match the conditioning labels.

For all tasks we use a GRU (Gated Recurrent Unit \cite{chung2015gated}) RNN with hidden state size 500 as the encoder $G_\text{enc}$.
Attribute labels are represented as a binary vector, and an attribute embedding is constructed via linear projection.
The decoder $G_\text{dec}$ is initialized using a concatenation of the representation coming from the encoder and the attribute embedding. 
Attribute embeddings of size 200 and a decoder GRU with hidden state size 700 were used (These parameters are identical to \cite{shen2017style}).
The discriminator receives an RNN hidden state sequence and an attribute vector as input. 
The hidden state sequence is encoded using a bi-directional RNN $\phi$ with hidden state size 500.
The interpolation probability $\Gamma \in \{0, 0.1, 0.2, .., 1.0\}$ and weight of the adversarial loss $\lambda \in \{0.5, 1.0, 1.5\}$ are chosen based on the validation metrics above.
Word embeddings are initialized with pre-trained GloVe embeddings \citep{pennington2014glove}.

\subsection{Metrics}
\label{sec:metrics}

Although the evaluation setups in prior work assess how well the generated sentences match the conditioning labels, they do not assess whether they match the input sentence in content. 
For most attributes of interest, parallel corpora do not exist. 
Hence we define objective metrics that evaluate models in a setting where ground truth annotations are unavailable.
While individually these metrics have their deficiencies, taken together they are helpful in objectively comparing different models and performing consistent evaluations across different work.

\textbf{Attribute accuracy} 
\, To quantitatively evaluate how well the generated samples match the conditioning labels we adopt a protocol similar to \cite{hu2017controllable}.
We generate samples from the model and measure label accuracy using a pre-trained sentiment classifier.
For the sentiment experiments, the pre-trained classifiers are CNNs trained to perform sentiment analysis at the review level on the Yelp  and IMDB datasets \citep{maas2011learning}.
The classifiers achieve test accuracies of 95\%, 90\% on the respective datasets.

\textbf{Content compatibility} 
\, Measuring content preservation using objective metrics is challenging. 
\citet{fu2017style} propose a content preservation metric which extracts features from word embeddings and measures cosine similarity in the feature space.
However, it is hard to design an embedding based metric which disregards the attribute information present in the sentence.
We take an indirect approach and measure properties that would hold if the models do indeed produce content compatible sentences.
We consider a content preservation metric inspired by the unsupervised model selection criteria of \citet{lample2017unsupervised} to evaluate machine-translation models without parallel data. 
Given two non-parallel datasets $D_\text{src}, D_\text{tgt}$ and translation models $M_{src\rightarrow tgt}, M'_{tgt\rightarrow src}$ that map between the two domains, the following metric is defined. 
\begin{equation}
f_\text{content}(M,M') = 
0.5  [\mathbb{E}_{x\sim D_\text{src}} \text{BLEU}(x, M'\circ M(x)) + \mathbb{E}_{x\sim D_\text{tgt}} \text{BLEU}(x, M\circ M'(x))]
\label{MS}
\end{equation}
where $M\circ M' (x)$ represents translating $x\in D_\text{src}$ to domain $D_\text{tgt}$ and then back to $D_\text{src}$.
We assume $D_\text{src}$ and $D_\text{tgt}$ to be test set sentences of positive and negative sentiment respectively and $M,M'$ to be the generative model conditioned on positive and negative sentiment, respectively.

\vspace{-1.0em}
\textbf{Fluency}
\, We use a pre-trained language model to measure the fluency of generated sentences.
The perplexity of generated sentences, as evaluated by the language model, is treated as a measure of fluency.
A state-of-the-art language model trained on the Billion words benchmark \citep{jozefowicz2016exploring} dataset is used for the evaluation.




\begin{table*}
\centering
\setlength\tabcolsep{5.6pt}
\begin{tabular}{l | c  c  c  c| c  c  c  c}
\hline
& \multicolumn{4}{c|}{Yelp Reviews} & \multicolumn{4}{c}{IMDB Reviews} \\
\hline
Model & Attribute $\uparrow$ & \multicolumn{2}{c}{Content $\uparrow$} & Fluency $\downarrow$ & Attribute $\uparrow$ & \multicolumn{2}{c}{Content $\uparrow$} & Fluency $\downarrow$ \\
& Accuracy & B-1 & B-4 & Perp. & Accuracy & B-1 & B-4 & Perp. \\
\hline
Ctrl-gen \citep{hu2017controllable}      & 76.36\% & 11.5 & 0.0 & 156 & 76.99\% & 15.4 & 0.1 & 94 \\
Cross-align \citep{shen2017style}        & 90.09\% & 41.9 & 3.9 & 180 & 88.68\% & 31.1 & 1.1 & 63 \\
Ours                                     & \textbf{90.50\%} & \textbf{53.0} & \textbf{7.5} & \textbf{133} & \textbf{94.46\%} & \textbf{40.3} & \textbf{2.2} & \textbf{52} \\
\hline
\end{tabular}
\caption{Quantitative evaluation for sentiment conditioned generation. Attribute compatibility represents label accuracy of generated sentences, as measured by a pre-trained classifier. Content preservation is assessed based on $f_\text{content}$ (BLEU-1 (B-1) and BLEU-4 (B-4) scores). Fluency is evaluated in terms of perplexity of generated sentences as measured by a pre-trained classifier. Higher numbers are better for accuracy and content compatibility, and lower numbers are better for perplexity.}
\label{metrics}
\end{table*}
\cuttablebelow

\begin{table}[!t]
\RawFloats
\setlength{\tabcolsep}{3pt} 
\begin{minipage}{.48\linewidth}
\centering
\begin{tabular}{l | c  c  c}
\hline
Model & Attribute & Content & Fluency \\
\hline
Ctrl-gen \citep{hu2017controllable}      & 66.0\% & 6.94\% & 2.51 \\
Cross-align \citep{shen2017style}   & 91.2\% & 22.04\% & 2.54 \\
Ours           & \textbf{92.8\%} & \textbf{55.10\%} & \textbf{3.19} \\
\hline               
\end{tabular}
\caption{Human assessment of sentences generated by the models. Attribute and content scores indicate percentage of sentences judged by humans to have the appropriate attribute label and content respectively. Fluency scores were obtained on a 5 point Likert scale. }
\label{human}
\end{minipage}
\hfill
\begin{minipage}{.49\linewidth}
\centering
\begin{tabular}{l | l | c}
\hline
Supervision & Model & BLEU \\
\hline
\multirow{2}{*}{Paired data} & Seq2seq & 10.4 \\
& Seq2seq-bi & 11.15  \\
\hline
Unpaired data & Ours & 7.65 \\
Paired + Unpaired data & Ours & \textbf{13.89} \\
\hline
\end{tabular}
\caption{Translating Old English to Modern English. The seq2seq models are supervised with parallel data. We consider our model in the unsupervised (no parallel data) and semi-supervised (paired and unpaired data) settings.}
\label{shakespeare_bleu}
\end{minipage}
\cuttablebelow
\end{table}

\subsection{Sentiment Experiments}
\vspace{-0.1em}
\textbf{Data} \, We use the restaurant reviews dataset from \cite{shen2017style}.
The dataset is a filtered version of the Yelp reviews dataset. 
Similar to \cite{hu2017controllable}, we use the IMDB move review corpus from \cite{diao2014jointly}.
We use \citet{shen2017style}'s filtering process to construct a subset of the data for training and testing.
The datasets respectively have 447k, 300k training sentences and 128k, 36k test sentences. 



We compare our model against Ctrl-gen, the VAE model of \citet{hu2017controllable} and Cross-align, the cross alignment model of \citet{shen2017style}.
Code obtained from the authors is used to train models on the datasets.
We use a pre-trained model provided by \cite{hu2017controllable} for movie review experiments.

\textbf{Quantitative evaluation}
\, Table \ref{metrics} compares our model against prior work in terms of the objective metrics discussed in the previous section.
Both \cite{shen2017style,hu2017controllable} perform soft-decoding, so that back-propagation through the sampling process is made possible.
But this leads to artifacts in generation, producing low fluency scores.
Note that the fluency scores do not represent the perplexity of the generators, but perplexity measured on generated sentences using a pre-trained language model.
While the absolute numbers may not be representative of the generation quality, it serves as a useful measure for relative comparison.

We report BLEU-1 and BLEU-4 scores for the content metric.
Back-translation has been effectively used for data augmentation in unsupervised translation approaches.
The interpolation loss can be thought of as data augmentation in the feature space, taking into account the noisy nature of parallel text produced by the model, and encourages content preservation when modifying attribute properties. 
The cross-align model performs strongly in terms of attribute accuracy, however it has difficulties generating grammatical text. 
Our model is able to outperform these methods in terms of all metrics.

\textbf{Qualitative evaluation}
\, Table \ref{comp} shows samples generated from the models for given conditioning sentence and sentiment label.
For each query sentence, we generate a sentence conditioned on the opposite label.
The Ctrl-gen model rarely produces content compatible sentences. 
Cross-align produces relevant sentences, while parts of the sentence are ungrammatical.
Our model generates sentences that are more related to the input sentence.
More examples can be found in the supplementary material.

\newpage
\begin{table*}[h]
\begin{tabular}{l|p{0.77\textwidth}}
\hline
\multicolumn{2}{c}{\textbf{Restaurant reviews}} \\
\hline
\multicolumn{2}{c}{negative $\rightarrow$ positive} \\
\hline
Query & \textit{the people behind the counter were not friendly whatsoever .} \\
Ctrl gen \citep{hu2017controllable} & the food did n't taste as fresh as it could have been either . \\
Cross-align \citep{shen2017style} & the owners are the staff is so friendly . \\
Ours & the people at the counter were very friendly and helpful . \\ 
\hline
\multicolumn{2}{c}{positive $\rightarrow$ negative} \\
\hline
Query & \textit{they do an exceptional job here , the entire staff is professional and accommodating !} \\
Ctrl gen \citep{hu2017controllable} & very little water just boring ruined ! \\
Cross-align \citep{shen2017style} & they do not be back here , the service is so rude and do n't care ! \\
Ours & they do not care about customer service , the staff is rude and unprofessional ! \\
\hline
\multicolumn{2}{c}{\textbf{Movie reviews}} \\
\hline
\multicolumn{2}{c}{negative $\rightarrow$ positive} \\
\hline
Query & \textit{once again , in this short , there isn't much plot .} \\
Ctrl gen \citep{hu2017controllable} & it's perfectly executed with some idiotically amazing directing .\\
Cross-align \citep{shen2017style} & but <unk> , , the film is so good , it is . \\
Ours & first off , in this film , there is nothing more interesting . \\
\hline
\multicolumn{2}{c}{positive $\rightarrow$ negative} \\
\hline
Query & \textit{that's another interesting aspect about the film .} \\
Ctrl gen \citep{hu2017controllable} & peter was an ordinary guy and had problems we all could <unk> with \\
Cross-align \citep{shen2017style} & it's the <unk> and the plot . \\
Ours & there's no redeeming qualities about the film . \\
\hline
\end{tabular}
\caption{Query sentences modified with opposite sentiment by Ctrl gen \citep{hu2017controllable}, Cross-align \citep{shen2017style} and our model, respectively.}
\label{comp}
\end{table*}

\textbf{Human evaluation}
\, We supplement the quantitative and qualitative evaluations with human assessments of generated sentences.
Human judges on MTurk were asked to rate the three aspects of generated sentences we are interested in - attribute compatibility, content preservation and fluency.
We chose 100 sentences from the test set randomly and generated corresponding sentences with the same content and opposite sentiment.
Attribute compatibility is assessed by asking judges to label generated sentences and comparing the opinions with the actual conditioning sentiment label. 
For content assessment, we ask judges whether the original and generated sentences are related by the desired property (same semantic content and opposite sentiment).
Fluency/grammaticality ratings were obtained on a 5-point Likert scale. 
More details about the evaluation setup are provided in section \ref{sec:human} of the appendix.
Results are presented in Table \ref{human}.
These ratings are in agreement with the objective evaluations and indicate that samples from our model are more realistic and reflect the conditioning information better than previous methods. 

\vspace*{-0.5em}\subsection{Monolingual Translation}
We next consider a style transfer experiment where we attempt to emulate a particular writing style.
This has been traditionally formulated as a monolingual translation problem where aligned data from two styles are used to train translation models.
We consider English texts written in old English and address the problem of translating between old and modern English.
We used a dataset of Shakespeare plays crawled from the web \citep{xu2012paraphrasing}.
A subset of the data has alignments between the two writing styles. 
The aligned data was split as 17k pairs for training and 2k, 1k pairs respectively for development and test.
All remaining ~80k sentences are considered unpaired. 

We consider two sequence to sequence models as baselines.
The first one is a simple sequence to sequence model that is trained to translate old to modern English.
The second variation learns to translate both ways, where the decoder takes the domain of the target sentence as an additional input.
We compare the performance of models in Table \ref{shakespeare_bleu}.
In addition to the unsupervised setting which doesn't use any parallel data, we also train our model in the semi-supervised setting.
In this setting we first train the model using supervised sequence-to-sequence learning and fine-tune on the unpaired data using our objective.
Our version of the model that does not use any aligned data falls short of the supervised models.
However, in the semi-supervised setting we observe an improvement of more than 2 BLEU points over the purely supervised baselines.
This shows that the model is capable of finding sentence alignments by exploiting the unlabelled data.

\newpage
\begin{table}[h]
\begin{tabular}{c  c  c  c l}
\hline
\textbf{Mood} & \textbf{Tense} & \textbf{Voice} & \textbf{Neg.} & \textbf{john was born in the camp}\\
\hline
Indicative  & Past      & Passive   & No    & john was born in the camp . \\
Indicative  & Past      & Passive   & Yes   & john wasn't born in the camp . \\
Indicative  & Past      & Active    & No    & john had lived in the camp . \\
Indicative  & Past      & Active    & Yes   & john didn't live in the camp . \\
Indicative  & Present   & Passive   & No    & john is born in the camp . \\
Indicative  & Present   & Passive   & Yes   & john isn't born in the camp . \\
Indicative  & Present   & Active    & No    & john has lived in the camp . \\
Indicative  & Present   & Active    & Yes   & john doesn't live in the camp . \\
Indicative  & Future    & Passive   & No    & john will be born in the camp . \\
Indicative  & Future    & Passive   & Yes   & john will not be born in the camp . \\
Indicative  & Future    & Active    & No    & john will live in the camp . \\
Indicative  & Future    & Active    & Yes   & john will not survive in the camp . \\
Subjunctive & Cond      & Passive   & No    & john could be born in the camp . \\
Subjunctive & Cond      & Passive   & Yes   & john couldn't live in the camp . \\
Subjunctive & Cond      & Active    & No    & john could live in the camp . \\
Subjunctive & Cond      & Active    & Yes   & john couldn't live in the camp . \\
\hline
\end{tabular}
\caption{Simultaneous control of multiple attributes. Generated sentences for all valid combinations of the input attribute values.}
\label{simul}
\end{table}

\vspace*{-1.5em}
\subsection{Ablative study}
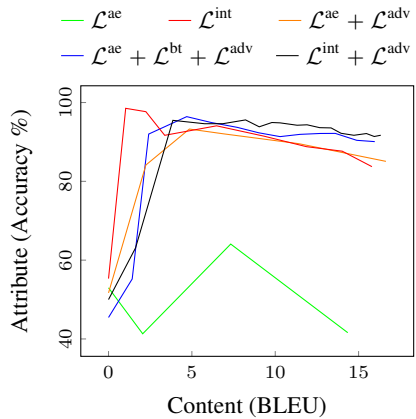
\begin{wrapfigure}{r}{0.4\textwidth}
\vspace{-2.0em}
\begin{tikzpicture}
\pgfplotstableread[col sep=comma]{Ablation/aebt.csv}\aebt
\pgfplotstableread[col sep=comma]{Ablation/ours.csv}\intcgan
\pgfplotstableread[col sep=comma]{Ablation/aecgan.csv}\aecgan
\pgfplotstableread[col sep=comma]{Ablation/int.csv}\int
\pgfplotstableread[col sep=comma]{Ablation/ae.csv}\ae
\begin{axis}[
xlabel=Content (BLEU),
ylabel=Attribute (Accuracy \%), 
xtick pos=left,
ytick pos=left,
width=0.43\textwidth, 
xticklabel style = {font=\scriptsize}, 
yticklabel style = {font=\scriptsize,rotate=90}, 
major tick length=2pt, 
x label style={font=\small, at={(axis description cs:0.5,0.03)},anchor=north},
y label style={font=\small, at={(axis description cs:0.16,.5)},anchor=south},
legend style={at={(1.30,1.30)}, draw=none, fill=none, font=\scriptsize, mark=square, line width=1pt},
legend cell align={right},
legend image post style={scale=0.5}
]
\addplot[mark=*, mark options={scale=0.0}, color=green] table[x = content, y = attribute] from \ae;
\label{p1}
\addplot[mark=*, mark options={scale=0.0}, color=blue] table[x = content, y = attribute] from \aebt;
\label{p2}
\addplot[mark=triangle*, mark options={scale=0.0}, color=orange] table[x = content, y = attribute] from \aecgan;
\label{p3}
\addplot[mark=diamond*, mark options={scale=0.0}, color=red] table[x = content, y = attribute] from \int;
\label{p4}
\addplot[mark=square*, mark options={scale=0.0}, color=black]  table[x = content, y = attribute] from \intcgan;
\label{p5}
\end{axis}                                                

\node [fill=white] at (rel axis cs: 0.55,1.33) {\shortstack[l]{
\ref*{p1} {\small $\mathcal{L}^\text{ae}$} \hspace{1.2em} \ref*{p4} {\small $\mathcal{L}^\text{int}$} \hspace{1.2em} \ref*{p3} {\small $\mathcal{L}^\text{ae}+\mathcal{L}^\text{adv}$}}};

\node [fill=white] at (rel axis cs: 0.55,1.2) {\shortstack[l]{
\ref*{p2} {\small $\mathcal{L}^\text{ae}+\mathcal{L}^\text{bt}+\mathcal{L}^\text{adv}$} \hspace{0.3em}
\ref*{p5} {\small $\mathcal{L}^\text{int}+\mathcal{L}^\text{adv}$}}};
\end{tikzpicture}

\caption{\small For different objective functions, we plot attribute compatibility against content compatibility of the learned model as training progresses. Models at the top right are desirable (High compatibility along both dimensions). }
\label{ablative}

\end{wrapfigure}

Figure \ref{ablative} shows an ablative study of the different loss components of the model. 
Each point in the plots represents the performance of a model (on the validation set) during training, where we plot the attribute compatibility against content compatibility.
As training progresses, models move to the right. 
Models at the top right are desirable (high attribute and content compatibility). 
$\mathcal{L}^\text{ae}$ and $\mathcal{L}^\text{int}$ refer to models trained with only the auto-encoding loss or the interpolated loss respectively.
We observe that the interpolated reconstruction loss by itself produces a reasonable model. 
It augments the data with generated samples and acts as a regularizer. 
Integrating the adversarial loss $\mathcal{L}^\text{adv}$ to each of the above losses improves the attribute compatibility since it explicitly requires generated sequences to be label compatible (and realistic). 
We also consider $\mathcal{L}^\text{ae} + \mathcal{L}^\text{bt} + \mathcal{L}^\text{adv}$ in our control experiment.
While this model performs strongly, it suffers from the issues associated with $\mathcal{L}^\text{ae}$ and $\mathcal{L}^\text{bt}$ discussed in section \ref{sec:content}.
The attribute compatibility of the proposed model $\mathcal{L}^\text{int} + \mathcal{L}^\text{adv}$ drops more gracefully compared to the other settings as the content preservation improves. 

\vspace*{-0.5em}
\subsection{Simultaneous control of multiple attributes}
In this section we discuss experiments on simultaneously controlling multiple attributes of the input sentence.
Given a set of sentences annotated with multiple attributes, our goal is to be able to plug this data into the learning algorithm and obtain a model capable of tweaking these properties of a sentence. 
Towards this end, we consider the following four attributes: tense, voice, mood and negation.
We use an annotation tool \citep{ramm2017annotating} to annotate a large corpus of sentences.
We do not make fine distinctions such as progressive and perfect tenses and collapse them into a single category.
We used a subset of $\sim$2M sentences from the BookCorpus dataset \citep{kiros2014multiplicative}, chosen to have approximately near class balance across different attributes.

Table \ref{simul} shows generated sentences conditioned on all valid combinations of attribute values for a given query sentence.
We use the annotation tool to assess attribute compatibility of generated sentences.
Attribute accuracies measured on generated senetences for mood, tense, voice, negation were respectively 98\%, 98\%, 90\%, 97\%.
The voice attribute is more difficult to control compared to the other attributes since some sentences require global changes such as switching the subject-verb-object order, and we found that the model tends to distort the content during voice control.  

\vspace{-0.5em}
\section{Conclusion}
\vspace{-0.5em}
In this work we considered the problem of modifying textual attributes in sentences.
We proposed a model that explicitly encourages content preservation, attribute compatibility and generating realistic sequences through carefully designed reconstruction and adversarial losses.
We demonstrate that our model effectively reflects the conditioning information through various experiments and metrics.
While previous work has been centered around controlling a single attribute and transferring between two styles, the proposed model easily extends to the multiple attribute scenario.
It would be interesting future work to consider attributes with continuous values in this framework and a much larger set of semantic and syntactic attributes.


\textbf{Acknowledgements}
We thank Andrew Dai, Quoc Le, Xinchen Yan and Ruben Villegas for helpful discussions. We also thank Jongwook Choi, Junhyuk Oh, Kibok Lee, Seunghoon Hong, Sungryull Sohn, Yijie Guo, Yunseok Jang and Yuting Zhang for helpful feedback on the manuscript. 

\bibliography{nips}

\begin{thebibliography}{34}
\providecommand{\natexlab}[1]{#1}
\providecommand{\url}[1]{\texttt{#1}}
\expandafter\ifx\csname urlstyle\endcsname\relax
  \providecommand{\doi}[1]{doi: #1}\else
  \providecommand{\doi}{doi: \begingroup \urlstyle{rm}\Url}\fi

\bibitem[Kingma and Welling(2013)]{kingma2013auto}
Diederik~P Kingma and Max Welling.
\newblock Auto-encoding variational bayes.
\newblock \emph{arXiv preprint arXiv:1312.6114}, 2013.

\bibitem[Goodfellow et~al.(2014)Goodfellow, Pouget-Abadie, Mirza, Xu,
  Warde-Farley, Ozair, Courville, and Bengio]{goodfellow2014generative}
Ian Goodfellow, Jean Pouget-Abadie, Mehdi Mirza, Bing Xu, David Warde-Farley,
  Sherjil Ozair, Aaron Courville, and Yoshua Bengio.
\newblock Generative adversarial nets.
\newblock In \emph{Advances in neural information processing systems}, pages
  2672--2680, 2014.

\bibitem[van~den Oord et~al.(2016)van~den Oord, Kalchbrenner, and
  Kavukcuoglu]{oord2016pixel}
Aaron van~den Oord, Nal Kalchbrenner, and Koray Kavukcuoglu.
\newblock Pixel recurrent neural networks.
\newblock \emph{arXiv preprint arXiv:1601.06759}, 2016.

\bibitem[Yan et~al.(2016)Yan, Yang, Sohn, and Lee]{yan2016attribute2image}
Xinchen Yan, Jimei Yang, Kihyuk Sohn, and Honglak Lee.
\newblock Attribute2image: Conditional image generation from visual attributes.
\newblock In \emph{European Conference on Computer Vision}, pages 776--791.
  Springer, 2016.

\bibitem[Reed et~al.(2016)Reed, Akata, Yan, Logeswaran, Schiele, and
  Lee]{reed2016generative}
Scott Reed, Zeynep Akata, Xinchen Yan, Lajanugen Logeswaran, Bernt Schiele, and
  Honglak Lee.
\newblock Generative adversarial text to image synthesis.
\newblock \emph{arXiv preprint arXiv:1605.05396}, 2016.

\bibitem[Gatys et~al.(2015)Gatys, Ecker, and Bethge]{gatys2015neural}
Leon~A Gatys, Alexander~S Ecker, and Matthias Bethge.
\newblock A neural algorithm of artistic style.
\newblock \emph{arXiv preprint arXiv:1508.06576}, 2015.

\bibitem[Isola et~al.(2016)Isola, Zhu, Zhou, and Efros]{isola2016image}
Phillip Isola, Jun-Yan Zhu, Tinghui Zhou, and Alexei~A Efros.
\newblock Image-to-image translation with conditional adversarial networks.
\newblock \emph{arXiv preprint arXiv:1611.07004}, 2016.

\bibitem[Li et~al.(2016)Li, Galley, Brockett, Spithourakis, Gao, and
  Dolan]{li2016persona}
Jiwei Li, Michel Galley, Chris Brockett, Georgios~P Spithourakis, Jianfeng Gao,
  and Bill Dolan.
\newblock A persona-based neural conversation model.
\newblock \emph{arXiv preprint arXiv:1603.06155}, 2016.

\bibitem[Xu et~al.(2012)Xu, Ritter, Dolan, Grishman, and
  Cherry]{xu2012paraphrasing}
Wei Xu, Alan Ritter, William~B Dolan, Ralph Grishman, and Colin Cherry.
\newblock Paraphrasing for style.
\newblock In \emph{24th International Conference on Computational Linguistics,
  COLING 2012}, 2012.

\bibitem[Sennrich et~al.(2016)Sennrich, Haddow, and
  Birch]{sennrich2016controlling}
Rico Sennrich, Barry Haddow, and Alexandra Birch.
\newblock Controlling politeness in neural machine translation via side
  constraints.
\newblock In \emph{Proceedings of NAACL-HLT}, pages 35--40, 2016.

\bibitem[Shetty et~al.(2017)Shetty, Schiele, and Fritz]{shetty2017author}
Rakshith Shetty, Bernt Schiele, and Mario Fritz.
\newblock Author attribute anonymity by adversarial training of neural machine
  translation.
\newblock \emph{arXiv preprint arXiv:1711.01921}, 2017.

\bibitem[Sutskever et~al.(2014)Sutskever, Vinyals, and
  Le]{sutskever2014sequence}
Ilya Sutskever, Oriol Vinyals, and Quoc~V Le.
\newblock Sequence to sequence learning with neural networks.
\newblock In \emph{Advances in neural information processing systems}, pages
  3104--3112, 2014.

\bibitem[Kikuchi et~al.(2016)Kikuchi, Neubig, Sasano, Takamura, and
  Okumura]{kikuchi2016controlling}
Yuta Kikuchi, Graham Neubig, Ryohei Sasano, Hiroya Takamura, and Manabu
  Okumura.
\newblock Controlling output length in neural encoder-decoders.
\newblock \emph{arXiv preprint arXiv:1609.09552}, 2016.

\bibitem[Yamagishi et~al.(2016)Yamagishi, Kanouchi, Sato, and
  Komachi]{yamagishi2016controlling}
Hayahide Yamagishi, Shin Kanouchi, Takayuki Sato, and Mamoru Komachi.
\newblock Controlling the voice of a sentence in japanese-to-english neural
  machine translation.
\newblock In \emph{Proceedings of the 3rd Workshop on Asian Translation
  (WAT2016)}, pages 203--210, 2016.

\bibitem[Kiros et~al.(2014)Kiros, Zemel, and
  Salakhutdinov]{kiros2014multiplicative}
Ryan Kiros, Richard Zemel, and Ruslan~R Salakhutdinov.
\newblock A multiplicative model for learning distributed text-based attribute
  representations.
\newblock In \emph{Advances in neural information processing systems}, pages
  2348--2356, 2014.

\bibitem[Radford et~al.(2017)Radford, Jozefowicz, and
  Sutskever]{radford2017learning}
Alec Radford, Rafal Jozefowicz, and Ilya Sutskever.
\newblock Learning to generate reviews and discovering sentiment.
\newblock \emph{arXiv preprint arXiv:1704.01444}, 2017.

\bibitem[Hochreiter and Schmidhuber(1997)]{hochreiter1997long}
Sepp Hochreiter and J{\"u}rgen Schmidhuber.
\newblock Long short-term memory.
\newblock \emph{Neural computation}, 9\penalty0 (8):\penalty0 1735--1780, 1997.

\bibitem[Hu et~al.(2017)Hu, Yang, Liang, Salakhutdinov, and
  Xing]{hu2017controllable}
Zhiting Hu, Zichao Yang, Xiaodan Liang, Ruslan Salakhutdinov, and Eric~P Xing.
\newblock Controllable text generation.
\newblock \emph{arXiv preprint arXiv:1703.00955}, 2017.

\bibitem[Bowman et~al.(2015)Bowman, Vilnis, Vinyals, Dai, Jozefowicz, and
  Bengio]{bowman2015generating}
Samuel~R Bowman, Luke Vilnis, Oriol Vinyals, Andrew~M Dai, Rafal Jozefowicz,
  and Samy Bengio.
\newblock Generating sentences from a continuous space.
\newblock \emph{arXiv preprint arXiv:1511.06349}, 2015.

\bibitem[Chen et~al.(2016)Chen, Kingma, Salimans, Duan, Dhariwal, Schulman,
  Sutskever, and Abbeel]{chen2016variational}
Xi~Chen, Diederik~P Kingma, Tim Salimans, Yan Duan, Prafulla Dhariwal, John
  Schulman, Ilya Sutskever, and Pieter Abbeel.
\newblock Variational lossy autoencoder.
\newblock \emph{arXiv preprint arXiv:1611.02731}, 2016.

\bibitem[Li et~al.(2018)Li, Jia, He, and Liang]{li2018delete}
Juncen Li, Robin Jia, He~He, and Percy Liang.
\newblock Delete, retrieve, generate: A simple approach to sentiment and style
  transfer.
\newblock \emph{arXiv preprint arXiv:1804.06437}, 2018.

\bibitem[Shen et~al.(2017)Shen, Lei, Barzilay, and Jaakkola]{shen2017style}
Tianxiao Shen, Tao Lei, Regina Barzilay, and Tommi Jaakkola.
\newblock Style transfer from non-parallel text by cross-alignment.
\newblock \emph{arXiv preprint arXiv:1705.09655}, 2017.

\bibitem[Prabhumoye et~al.(2018)Prabhumoye, Tsvetkov, Salakhutdinov, and
  Black]{prabhumoye2018style}
Shrimai Prabhumoye, Yulia Tsvetkov, Ruslan Salakhutdinov, and Alan~W Black.
\newblock Style transfer through back-translation.
\newblock \emph{arXiv preprint arXiv:1804.09000}, 2018.

\bibitem[He et~al.(2016)He, Xia, Qin, Wang, Yu, Liu, and Ma]{he2016dual}
Di~He, Yingce Xia, Tao Qin, Liwei Wang, Nenghai Yu, Tieyan Liu, and Wei-Ying
  Ma.
\newblock Dual learning for machine translation.
\newblock In \emph{Advances in Neural Information Processing Systems}, pages
  820--828, 2016.

\bibitem[Artetxe et~al.(2017)Artetxe, Labaka, Agirre, and
  Cho]{artetxe2017unsupervised}
Mikel Artetxe, Gorka Labaka, Eneko Agirre, and Kyunghyun Cho.
\newblock Unsupervised neural machine translation.
\newblock \emph{arXiv preprint arXiv:1710.11041}, 2017.

\bibitem[Lample et~al.(2017)Lample, Denoyer, and
  Ranzato]{lample2017unsupervised}
Guillaume Lample, Ludovic Denoyer, and Marc'Aurelio Ranzato.
\newblock Unsupervised machine translation using monolingual corpora only.
\newblock \emph{arXiv preprint arXiv:1711.00043}, 2017.

\bibitem[Miyato and Koyama(2018)]{miyato2018cgans}
Takeru Miyato and Masanori Koyama.
\newblock cgans with projection discriminator.
\newblock \emph{arXiv preprint arXiv:1802.05637}, 2018.

\bibitem[Chung et~al.(2015)Chung, Gulcehre, Cho, and Bengio]{chung2015gated}
Junyoung Chung, Caglar Gulcehre, Kyunghyun Cho, and Yoshua Bengio.
\newblock Gated feedback recurrent neural networks.
\newblock In \emph{International Conference on Machine Learning}, pages
  2067--2075, 2015.

\bibitem[Pennington et~al.(2014)Pennington, Socher, and
  Manning]{pennington2014glove}
Jeffrey Pennington, Richard Socher, and Christopher~D Manning.
\newblock Glove: Global vectors for word representation.
\newblock In \emph{EMNLP}, volume~14, pages 1532--1543, 2014.

\bibitem[Maas et~al.(2011)Maas, Daly, Pham, Huang, Ng, and
  Potts]{maas2011learning}
Andrew~L. Maas, Raymond~E. Daly, Peter~T. Pham, Dan Huang, Andrew~Y. Ng, and
  Christopher Potts.
\newblock Learning word vectors for sentiment analysis.
\newblock In \emph{Proceedings of the 49th Annual Meeting of the Association
  for Computational Linguistics: Human Language Technologies}, pages 142--150,
  Portland, Oregon, USA, June 2011. Association for Computational Linguistics.
\newblock URL \url{http://www.aclweb.org/anthology/P11-1015}.

\bibitem[Fu et~al.(2017)Fu, Tan, Peng, Zhao, and Yan]{fu2017style}
Zhenxin Fu, Xiaoye Tan, Nanyun Peng, Dongyan Zhao, and Rui Yan.
\newblock Style transfer in text: Exploration and evaluation.
\newblock \emph{arXiv preprint arXiv:1711.06861}, 2017.

\bibitem[Jozefowicz et~al.(2016)Jozefowicz, Vinyals, Schuster, Shazeer, and
  Wu]{jozefowicz2016exploring}
Rafal Jozefowicz, Oriol Vinyals, Mike Schuster, Noam Shazeer, and Yonghui Wu.
\newblock Exploring the limits of language modeling.
\newblock \emph{arXiv preprint arXiv:1602.02410}, 2016.

\bibitem[Diao et~al.(2014)Diao, Qiu, Wu, Smola, Jiang, and
  Wang]{diao2014jointly}
Qiming Diao, Minghui Qiu, Chao-Yuan Wu, Alexander~J Smola, Jing Jiang, and
  Chong Wang.
\newblock Jointly modeling aspects, ratings and sentiments for movie
  recommendation (jmars).
\newblock In \emph{Proceedings of the 20th ACM SIGKDD international conference
  on Knowledge discovery and data mining}, pages 193--202. ACM, 2014.

\bibitem[Ramm et~al.(2017)Ramm, Lo{\'a}iciga, Friedrich, and
  Fraser]{ramm2017annotating}
Anita Ramm, Sharid Lo{\'a}iciga, Annemarie Friedrich, and Alexander Fraser.
\newblock Annotating tense, mood and voice for english, french and german.
\newblock \emph{Proceedings of ACL 2017, System Demonstrations}, pages 1--6,
  2017.

\end{thebibliography}
\bibliographystyle{unsrtnat}

\newpage
\appendix

\section{Qualitative comparison}
\label{sec:supplemental}


Table \ref{yelp} shows samples from different models for given query sentences from the restaurant reviews test dataset and opposite sentiment.

\begin{table*}[!ht]
\begin{tabular}{l|p{0.82\textwidth}}
\hline
\multicolumn{2}{c}{\textbf{Restaurant Reviews}} \\
\hline
\multicolumn{2}{c}{negative $\rightarrow$ positive } \\
\hline
Query & \textit{sorry but i do n't get the rave reviews for this place .} \\
Ctrl gen & i ordered the nachos , have perfect seasonal beans on amazing amazing .  \\
Cross-align & sorry , i do n't be the best experience ever . \\
Ours & thanks but i love this place for lunch .  \\
\hline
Query & \textit{however my recent visit there made me change my mind entirely .} \\
Ctrl-gen & not like other target stores . \\
Cross-align & best little one time to go for in charlotte . \\
Ours & overall my experience here was great as well . \\
\hline
Query & \textit{okay so this place has been a pain even after i already moved out .} \\
Ctrl gen & like i mentioned , i thought to this these fun . \\
Cross-align & food and this place has been a good place to be back . \\
Ours & overall this is a great place to go when i 'm in town . \\
\hline
Query & \textit{personally i 'd rather spend my money at a business that appreciates my business .} \\
Ctrl gen & i became quite a gem at the beginning but we amazing fantastic amazing . \\
Cross align & then i will be back my time to get a regular time . \\
Ours & definitely i 'll definitely be back for a good haircut . \\
\hline
Query & \textit{seems their broth just has no flavor kick to it .} \\
Ctrl gen & i expected more for the price i paid . \\
Cross align & loved their menu , has a great place . \\
Ours & definitely it 's cooked perfectly and it 's delicious . \\
\hline
\multicolumn{2}{c}{positive $\rightarrow$ negative } \\
\hline
Query & \textit{best chinese food i 've had in a long time .} \\
Ctrl gen & very lousy texture and ruined . \\
Cross align & worst chinese food i 've had in a long in years . \\
Ours & worst food i 've had in a long time . \\
\hline
Query & \textit{high quality food at prices comparable to lower quality take out .} \\
Ctrl gen & the rock becomes my daughter ruined and it was terrible lousy lousy lousy \\
Cross align & terrible quality , , <unk> <unk> , \_num\_ \% of \$ \_num\_ minutes . \\
Ours & poor quality of food quality at all costs . \\
\hline
Query & \textit{my appetizer was also very good and unique .} \\
Ctrl gen & both were ruined . ruined \\
Cross align & my wife was just very bland and no flavor . \\
Ours & my chicken was very dry and had no flavor . \\
\hline
Query & \textit{everything tasted great and the service was excellent .} \\ 
Ctrl gen & but the real pleasure is the service department . \\
Cross align & everything tasted horrible and the service was very bad . \\
Ours & everything tasted bad and the service was horrible . \\
\hline
Query & \textit{atmosphere is cozy and comfortable .} \\ 
Ctrl gen & atmosphere is not good . \\
Cross align & rude is dirty and way in . \\
Ours & restaurant is dirty and dirty . \\
\hline
\end{tabular}
\caption{Query sentences modified with opposite sentiment by Ctrl gen \cite{hu2017controllable}, Cross-align \cite{shen2017style} and our model, respectively.}
\label{yelp}
\end{table*}

\newpage
Table \ref{imdb} shows samples from different models for given query sentences from the movie reviews test dataset and opposite sentiment.

\begin{table*}[!ht]
\begin{tabular}{l|p{0.82\textwidth}}
\hline
\multicolumn{2}{c}{\textbf{Movie Reviews}} \\
\hline
\multicolumn{2}{c}{negative $\rightarrow$ positive } \\
\hline
Query & \textit{this is the most vapid movie i have ever seen .} \\
Ctrl gen & if this grabs your interest , you may want to give it a try \\
Cross-align & this is a great movie that is so good . \\
Ours & this is the most beautiful movie i have ever seen . \\
\hline
Query & \textit{this 1944 film is too awful as it 's just incredible .} \\
Ctrl gen & <unk> the three dead world and <unk> 's <unk> is a cult in a life \\
Cross-align & this film is one of the best movies ever made . \\
Ours & this film is an excellent and it is definitely worth it . \\
\hline
Query & \textit{1 out of 10 .} \\
Ctrl gen & he 's cold and hateful exactly what his part <unk> \\
Cross-align & my rating of the cast . \\
Ours & 10 out of 10 . \\
\hline
Query & \textit{i always thought she was a colorless , plain jane .} \\
Ctrl gen & a great comedy all wrapped up in a tiny package ! \\
Cross-align & i think that is the best of the film . \\
Ours & i also thought she was a beautiful , talented actor . \\
\hline
Query & \textit{her character is truly hateful and her acting , if you can call it that , is strictly wretched .} \\
Ctrl gen & a great ` proper ' summer movie \\
Cross-align & <unk> , is the <unk> , and you can be able to be more than it to be . \\
Ours & his character is very funny , and in fact , it 's just what he does n't disappoint . \\
\hline
\multicolumn{2}{c}{positive $\rightarrow$ negative } \\
\hline
Query & \textit{this is one of his best efforts .} \\
Ctrl gen & as david <unk> picked up the franchise , it has just <unk> to pieces \\
Cross-align & this is a complete waste of time . \\
Ours & this is one of the worst films . \\
\hline
Query & \textit{if you love silent films , you 'll adore this one .} \\
Ctrl gen & nice photographic effects as jessica <unk> the process \\
Cross-align & if you 're no , but it is not bad . \\
Ours & if you love horror movies , do n't see this one . \\
\hline
Query & \textit{and congratulations to kino for a superb video restoration .} \\
Ctrl gen & peter <unk> is not that she 's not gone bad movie\\
Cross-align & but then , it 's a waste of time . \\
Ours & and save your money on this piece of garbage . \\
\hline
Query & \textit{the characters are portrayed vividly and realistically .} \\
Ctrl gen & problem is , not enough good work went into this \\
Cross-align & the characters are <unk> and <unk> . \\
Ours & the characters are completely unsympathetic and annoying . \\
\hline
Query & \textit{there are some of the most stunning and grisly combat scenes ever filmed .} \\
Ctrl gen & unfortunately the only thing you see is <unk> \\
Cross-align & there is no a <unk> , and the <unk> , <unk> and <unk> . \\
Ours & there are some of the most boring and boring scenes ever made . \\
\hline
\end{tabular}
\caption{Query sentences modified with opposite sentiment by Ctrl gen \cite{hu2017controllable}, Cross-align \cite{shen2017style} and our model, respectively.}
\label{imdb}
\end{table*}

\newpage

\section{Human Evaluation}
\label{sec:human}

Sections \ref{content}, \ref{attribute}, \ref{fluency} describe the setup for human evaluations done using Amazon Mechanical Turk (AMT) to obtain annotations for generated sentences.

\subsection{Content compatibility}
\label{content} 
Given a reference sentence and a set of candidate sentences, pick the candidates that have the \textbf{same semantic content as the reference sentence but have the opposite sentiment} (i.e., mean the opposite). Select all that apply. If you think neither of the given sentences have this property, choose \textit{No preference} (This can happen when all the candidate sentences are either semantically irrelevant to the reference sentence or have the incorrect sentiment). 
 
Example: \\
Reference sentence: \textit{This is a great movie !} \\
You would pick sentences such as  \\
\hspace*{0.5em} \cmark \hspace{0.2em} \textit{This is not a good movie.} \\
\hspace*{0.5em} \cmark \hspace{0.2em} \textit{This is a bad movie.} \\
The following sentences do not fit the criteria because they are either semantically irrelevant to the reference sentence or have the incorrect sentiment. \\
\hspace*{0.5em} \xmark \hspace{0.2em} \textit{I did not like the salad.} \\
\hspace*{0.5em} \xmark \hspace{0.2em} \textit{This is a wonderful movie.}

\subsection{Attribute compatibility}
\label{attribute} 

Pick the best sentiment based on the following criterion.

\begin{tabular}{l l}
\textit{Sentiment} & \textit{Guidance} \\
Positive & Sentence conveys positive sentiment. Eg: "I really liked the food." \\
Negative & Sentence conveys negative sentiment. Eg: "This was the worst experience ever." \\
Neutral & Sentence does not carry any sentiment information.
\end{tabular}

\subsection{Fluency}
\label{fluency}

Rate the grammaticality/fluency of the sentence based on the following criterion.

\begin{tabular}{l l}
\textit{Fluency}	& \textit{Guidance} \\
5 & The sentence is grammatical and does not have any grammar errors. \\
4 & Sentence is mostly grammatical except for one/two mistakes. \\
3 & Parts of the sentence are grammatical and sentence is somewhat coherent, but there \\ 
& are glaring errors. \\
2 & Too many grammatical errors and sentence is incoherent. \\
1 & Sentence is completely ungrammatical.
\end{tabular}

\newpage
\section{Sampling strategy}
\label{sec:sampling}
In this section we compare soft and hard sampling during training.
For the soft-sampling model, we use an exponential decay temperature annealing schedule with an initial temperature of 1. 
The temperature decays until it reaches 0.01 and remains constant afterwards.
Other parameters of the model are identical to section \ref{hyperparameters}.
We use the Yelp dataset for this experiment.
Table \ref{sampling} compares the models with respect to the metrics in section \ref{sec:metrics}.

Models learned with soft-sampling produce sentences judged to be highly attribute compatible.
However, the content compatibility is considerably poor and generated sentences have lower fluency.
This supports our claim that the training and inference behavior are mismatched when soft-sampled sequences are used for training. 
\begin{table*}[!h]
\centering
\begin{tabular}{l | c  c  c  c}
\hline
Model & Attribute $\uparrow$ & \multicolumn{2}{c}{Content $\uparrow$} & Fluency $\downarrow$ \\
& compatibility & \multicolumn{2}{c}{compatibility} & \\
& & BLEU-1 & BLEU-4 & Perplexity \\
\hline
Hard-sampling         & 90.50\% & \textbf{53.0} & \textbf{7.5} & \textbf{133} \\
Soft-sampling         & \textbf{92.33}\% & 43.6 & 3.1 & 137 \\
\hline
\end{tabular}
\caption{Comparison between soft and hard sampling. Evaluation metrics are described in section \ref{sec:metrics}. Higher numbers are better for accuracy and content compatibility and lower numbers for perplexity.}
\label{sampling}
\end{table*}

\end{document}